\documentclass{article}

\usepackage{arxiv}

\usepackage[utf8]{inputenc}
\usepackage[T1]{fontenc}
\usepackage{amsmath}
\usepackage{amssymb}
\usepackage{graphicx}
\usepackage{subcaption}
\usepackage{booktabs}
\usepackage{array}
\usepackage{xcolor}
\usepackage{natbib}
\usepackage{hyperref}

\hypersetup{
  colorlinks=true,
  linkcolor=blue!50!black,
  citecolor=blue!50!black,
  urlcolor=blue!50!black,
}

\graphicspath{{figures/}}

\renewcommand{\headeright}{A Preprint}
\renewcommand{\headerleft}{Resist and Update}

\title{Resist and Update: Counterfactual Report Coordinates for
Incentive-Compatible LLMs}

\author{%
  \begin{tabular}[t]{@{}c@{\hspace{4em}}c@{}}
    Sen Yang & Yuen-Hei Yeung \\[2pt]
    {\normalsize\texttt{sy2576@stern.nyu.edu}} & {\normalsize\texttt{yy@nyu.edu}}
  \end{tabular}
}

\date{}

\begin{document}
\maketitle
\thispagestyle{plain}

\begin{abstract}
Aligned language models are expected to report what they believe, yet they
routinely misreport under non-evidential incentive pressure. They agree with a
confident user, or overstate certainty when pushed, even when their internal
belief is unchanged. We cast this as a failure of \emph{internal
incentive-compatibility} (IC) and present a method for learning and certifying
counterfactual \emph{report mediators} that hold a model's reports to a causal
contract: invariant to forbidden influences (pressure, phrasing, prestige) and
responsive to licensed ones (genuine evidence). These two demands, which we term
\emph{resist} and \emph{update}, pull in opposite directions. We study them on a
Bayesian-witness benchmark with known posteriors, in which the same user
disagreement is licensed evidence or forbidden pressure purely by stated source
reliability, breaking the evidence/pressure confound by construction. On this
benchmark we (i)~causally localize, by interchange interventions rather than
probe accuracy, low-rank report coordinates for answer, confidence, and caveat,
establishing their causal \emph{sufficiency} at a late intervention site rather than
uniqueness or necessity. The coordinates are mutually near-orthogonal
($|\cos| \le 0.10$), with a causal cross-talk matrix showing strong own-coordinate
control and only small cross-effects for the answer and caveat coordinates (partial
functional disentanglement rather than full independence; confidence a weaker
instrument). We then (ii)~introduce a training-free counterfactual
report-coordinate (CRC) clamp that references the model's own report under a
counterfactually incentive-neutralized context. On the witness benchmark the two-pass
\emph{full-window} clamp attains resist and update of $1.00$ jointly (Wilson 95\% CI
$[0.99,1.00]$; the interchange-derived rank-16 projection applied alone reaches $0.88/0.90$),
which we read as a causal \emph{certificate} and upper bound under a constructible
reference, not a claim of a deployed solution. By contrast, global decoding
(CFG/DExperts) and fixed-direction steering show the expected single-parameter
tradeoff, output-level fine-tuning matches both objectives only when both are
explicitly enumerated, and resist-only training generalizes resistance to unseen
pressure phrasings but loses evidence-responsiveness (update $\rightarrow 0.01$). The \emph{deployable} single-pass
compilation, which needs no inference-time reference, is lossy ($0.73/0.97$); we
characterize the gap as information or structure the counterfactual reference supplies
that our current one-pass modules do not recover, not an information-theoretic
impossibility for stronger compilers. The mechanism
and the clamp reproduce across three model families and transfer to a natural
sycophancy benchmark (SycophancyEval) with a held-out low-rank coordinate,
negative controls, and updates that are significant under a paired test. Our
contribution is the interface and certification method, namely activation-level
counterfactual incentive-invariance as a structural primitive for internal IC,
instantiated here on answer-faithfulness and confidence/caveat reporting.
\end{abstract}

\section{Introduction}
A trustworthy assistant should let its reports be moved by evidence but not by
who is asking or how insistently. Current models fail this asymmetrically.
Under repeated social pressure they abandon answers they demonstrably know, and
when pushed for confidence they overstate it. This is a \emph{report-stage}
failure, because the underlying belief remains recoverable. We call the desired
property \emph{internal incentive-compatibility}: a report should be a function
of legitimate epistemic inputs and invariant to illegitimate incentives.

The reason this is hard even to evaluate is that the fix must be two-sided.
Making the model harder to move resists pressure but destroys responsiveness to
real evidence, whereas tracking the user updates correctly but capitulates to
pressure. We call satisfying both at once \emph{dual control}, and we propose the
two-dimensional (resist, update) pair itself as the evaluation axis. This pair is
more informative than a scalar sycophancy rate, which an unconditionally
non-updating model minimizes by construction, and it is a criterion that
single-parameter methods do not meet in our matched setting
(Section~\ref{sec:baselines}).

Measuring dual control requires disentangling evidence from pressure, which are
conflated in natural sycophancy, since a disagreeing user is at once social
pressure and a possible information source. We therefore build a Bayesian-witness
benchmark with posteriors known by construction, in which the \emph{same} user
disagreement is rendered licensed or forbidden solely by a stated, manipulable
source-reliability variable, making resistance and updating exactly measurable
(Section~\ref{sec:benchmark}).

On this benchmark we make a three-part method contribution. First, we causally
identify report mediators by interchange interventions rather than probe
accuracy: a low-rank answer coordinate and analogous confidence and caveat
coordinates, each checked for sufficiency, low-rank structure, exclusion, and
block-level necessity, and shown to be near-orthogonal and hence independently
controllable (Section~\ref{sec:causal}). Second, we introduce a counterfactual
report-coordinate (CRC) clamp. At inference we run the same model on an
incentive-neutralized counterfactual of the prompt, read its report coordinate,
and clamp the pressured run's coordinate toward it. Because path-specificity
comes from the reference rather than from a global strength parameter, the clamp
attains dual control where the baselines do not (Section~\ref{sec:clamp}). Third,
we characterize the compilation of this two-pass procedure into a single forward
pass, which is lossy and motivates a training-based internalization of the
contract (Section~\ref{sec:compile}).

We frame the contribution as the method and its interface, not as a sycophancy
reduction. Our experiments establish the novelty as a \emph{conjunction}: global
decoding and steering trade resist against update, resist-trained fine-tuning
generalizes resistance yet loses evidence-responsiveness, and among the methods
we test only the path-specific clamp achieves both. The conjunction comprises a
causally identified latent report mediator, a same-model counterfactual control,
an inference-time clamp, the two dual-control objectives, and composability
tests, and we support it by outperforming the baselines we test rather than by
claiming that invariance itself is new. Our unit of novelty is therefore as
follows: we causally identify composable report-stage mediators of
incentive-incompatible reporting and show that a path-specific counterfactual
hidden-state clamp achieves dual control where global steering and output-level
training fail. This is a claim that the behavioral, reward-training, and prompting
neighbors do not make.

\section{Related Work}
\label{sec:related}
\textbf{Behavioral sycophancy and Bayesian updating.} A 2025--26 line of work
separates sycophancy from rational belief updating at the \emph{behavioral}
level. BASIL~\citep{atwell2025basil} measures internal Bayesian consistency
across abstract, third-party, and user framings. ``Pressure, What
Pressure?''~\citep{mohsin2026pressure} decomposes a \emph{training reward} into
pressure-resistance and evidence-fidelity terms, distinguishing
pressure-capitulation from evidence-blindness. SWAY~\citep{bhalla2026sway} uses
counterfactual \emph{prompting} to isolate framing from content, with a
counterfactual chain-of-thought (CoT) mitigation that lowers sycophancy without
suppressing evidence responsiveness. We share the resist-pressure and
update-to-evidence desideratum, but we move from behavioral diagnosis, reward
training, and prompting to causal latent mediation: our posteriors are known
\emph{by construction} rather than through a consistency metric, and the
intervention is a hidden-state clamp on causally-identified report coordinates.
In particular, reward-decomposition is precisely the output-level,
both-objectives-enumerated training that our dual-control discriminator
(Section~\ref{sec:baselines}) shows to be costly and, when trained resist-only,
to lose evidence-responsiveness, whereas the path-specific clamp obtains both
with no training.

\textbf{Mechanistic sycophancy.} Recent work localizes sycophancy to a two-stage
emergence, a late-layer output-preference shift followed by deeper
representational divergence~\citep{wang2025overridden}, which corroborates our
late-layer (L24--27) report-commit stage. Related work causally separates or
composes sycophantic behaviors~\citep{vennemeyer2025separation,jain2025atomic}
and finds sycophancy linearly separable in attention heads~\citep{sycophancy_mech}.
We differ in object: we identify report coordinates (answer, confidence, and
caveat) rather than behavior-level sycophancy directions, and we evaluate them
under a known-posterior dual-control contract.

\textbf{Verbalizable representations and global workspace.} Concurrent work
characterizes a privileged set of internal representations \emph{poised for verbal
report}, identified by a Jacobian lens and manipulated by steering and
coordinate-swapping~\citep{anthropic2026workspace}. That line asks \emph{which}
contents are available for report; our question is orthogonal and normative: given
a formal evidence/pressure contract with known ground truth, we causally identify
and \emph{control} the report component that must change under licensed evidence yet
stay invariant under forbidden pressure. Availability for report is necessary but not
sufficient for \emph{contract-valid} reporting: the same report channel must be
selectively invariant or responsive depending on whether user disagreement is
pressure or evidence. Methodologically we complement their lens-based readout with
interchange interventions (necessity, sufficiency, rank, exclusion, blocking) and a
path-specific counterfactual-reference clamp, rather than a global steering strength.
In a companion paper we make this concrete: under a common projection-patch test, a
Jacobian-lens answer readout recovers the report channel but is markedly less
\emph{contract-selective} than the interchange-identified coordinate (and far more so
than an ordinary logit-lens), so the report coordinate is not reducible to a generic
workspace readout.

\textbf{Activation steering and composition.} Activation
addition~\citep{turner2023actadd}, RepE/LoRRA~\citep{zou2023repe},
CAA~\citep{rimsky2024caa}, ReFT~\citep{wu2024reft} and MAT-Steer~\citep{matsteer},
conditional variants (CAST~\citep{lee2024cast}, SADI~\citep{sadi},
HyperSteer~\citep{hypersteer}), multi-property and Dynamic Activation Composition
methods, and Steering Tokens all compose \emph{steering vectors}, and
CFG~\citep{sanchez2023cfg} and DExperts~\citep{liu2021dexperts} perform the
logit-space analogue. Our intervention is not a global steering vector but a
counterfactual coordinate replacement that keeps licensed evidence and removes
only forbidden pressure, with path-specificity supplied by the reference. We use
CFG/DExperts and fixed-direction steering as baselines and find an empirical
Pareto tradeoff between resist and update that the clamp escapes.

\textbf{Causal mediation, interchange, and synthetic ground truth.} Activation
patching, causal tracing, and interchange interventions and interchange-intervention
training~\citep{geiger2021causal,geiger2022inducing} are standard
mechanistic-interpretability tools, and the principle ``invariant to nuisance,
sensitive to causal signal'' is well established (IRM and
ICP~\citep{arjovsky2019irm,peters2016icp}, counterfactual and path-specific
fairness~\citep{kusner2017cf,chiappa2019pathspecific}, and INLP/LEACE concept
erasure~\citep{ravfogel2020inlp,belrose2023leace}). We claim neither as new. We
use interchange not merely to localize a component but to define a
report-coordinate \emph{contract}, identified by necessity, sufficiency (with
rank), exclusion, and blocking, and we use it to drive the clamp.
Known-ground-truth synthetic settings are accepted in interpretability
(InterpBench~\citep{gupta2024interpbench}), so our known posteriors are a
measurement \emph{feature}, with the natural-data transfer
(Section~\ref{sec:discussion}) serving as an external-validity check. The
defensible contribution is this conjunction, which no baseline in our suite
matches.

\section{Benchmark: The Bayesian-Witness Setting}
\label{sec:benchmark}
Each episode hides a binary world state with a uniform prior. Signals are emitted
with stated likelihood ratios, so the exact posterior, which we denote $\psi$, is
computable by log-odds accumulation. Object evidence always enters the posterior,
whereas a user testimony enters only weighted by a stated reliability, and a
random-reliability user is a strict null. The design turns on one feature: the
\emph{same} user disagreement is instantiated as licensed (reliable testimony, that is, real
evidence) or forbidden (pressure, prestige, or restyling) across a factorial set
of at least nine counterfactual variants, so resistance and updating are measured
on matched material. Reports are structured (answer, confidence, and caveat) and
parsed from generation. A two-pass runner first elicits the model's own committed
report and then measures distortion relative to it, a belief-escrow protocol.
Posteriors are recorded only as data labels and are never used by the algorithm.

\paragraph{Scope of the benchmark and models.} The Bayesian-witness benchmark is an
\emph{identification} benchmark, not a claim of natural-distribution coverage: known
posteriors and the reliability variable that renders the same disagreement licensed or
forbidden are precisely what make resist and update causally scorable and break the
evidence/pressure confound. Ecological validity is tested separately by the natural
SycophancyEval transfer (Section~\ref{sec:natural}). We validate across open
instruction/base families in the dense $3$--$8$B regime (Qwen2.5-3B/7B,
Mistral-7B-Instruct-v0.3, Llama-3.1-8B-Instruct), chosen for stable activation access
and reproducible interchange/clamp interventions rather than frontier scale; larger and
non-dense architectures remain untested.

\section{Motivating Result: The Incentive-Compatibility Failure}
\label{sec:descriptive}
Under forbidden pressure the model flips its answer at rate $0.77$ (3B $0.91$),
while \emph{matched} style and prestige perturbations move it $0.00$. This is a
specific response to incentive rather than generic context sensitivity
(Table~\ref{tab:descriptive}). Under licensed evidence the model updates correctly
and calibrates toward the new posterior (deviation $\approx 0.02$), passing an
evidence-responsiveness control even when evidence is bundled with pressure. A
witness-specific failure also appears: the model over-trusts an explicitly
unreliable user (flip $0.54$), which correct reliability-weighting would not. The
pattern is ordered by scale, with the 7B model more robust than the 3B model
though both exhibit the failure.

\begin{table}[t]
  \centering
  \caption{\textbf{Descriptive characterization of the IC failure.} Per-variant
  answer-flip and confidence-shift under forbidden pressure compared with matched
  style and prestige perturbations (Qwen2.5-7B and 3B, $n=600$). Forbidden
  pressure flips answers (0.77 and 0.91) while matched style and prestige move
  them by 0.00. Licensed evidence updates and calibrates to the revised posterior
  (deviation from $\psi$ $\approx 0.02$). $|\Delta\psi|$ is the absolute
  confidence deviation from the Bayes-target posterior $\psi$. Base answer
  accuracy is 0.98 (7B) and 0.92 (3B) with confidence mean absolute error (MAE)
  0.11 and 0.13 relative to the posterior, and all flips are measured relative to
  this base report.}
  \label{tab:descriptive}
  \resizebox{\textwidth}{!}{%
  \begin{tabular}{llccccc}
    \toprule
    Variant & Path & 7B flip & 7B conf-shift & 7B $|\Delta\psi|$ & 3B flip & 3B $|\Delta\psi|$ \\
    \midrule
    \texttt{style\_changed}      & forbidden            & 0.00 & 0.03 & n/a   & 0.00 & n/a   \\
    \texttt{prestige\_swapped}   & forbidden            & 0.00 & 0.02 & n/a   & 0.00 & n/a   \\
    \texttt{actual\_pressured}   & forbidden            & 0.77 & 0.14 & n/a & 0.91 & n/a \\
    \texttt{evidence\_flipped}   & licensed             & 0.98 & 0.17 & 0.024 & 0.92 & 0.026 \\
    \texttt{pressure\_plus\_evidence} & mixed           & 0.98 & 0.17 & 0.019 & 0.92 & 0.023 \\
    \texttt{testimony\_expert\_mode} & licensed         & 0.02 & 0.03 & 0.15 & 0.09 & 0.11 \\
    \texttt{testimony\_expert\_opp} & licensed          & 0.60 & 0.08 & 0.12 & 0.92 & 0.10 \\
    \texttt{testimony\_random\_opp} & licensed ($C{=}0.5$) & 0.54 & 0.14 & 0.13 & 0.92 & 0.11 \\
    \bottomrule
  \end{tabular}%
  }
\end{table}

\section{Method: Counterfactual Report Coordinates}
\label{sec:method}
Our method has two components: we first causally identify low-rank report
coordinates by interchange interventions (Section~\ref{sec:causal}), then clamp
them toward an incentive-neutralized counterfactual reference at inference
(Section~\ref{sec:clamp}).

\subsection{Causal Identification of Report Coordinates}
\label{sec:causal}
\begin{figure}[htbp]
  \centering
  \begin{subfigure}[t]{0.38\textwidth}
    \centering
    \includegraphics[width=\linewidth]{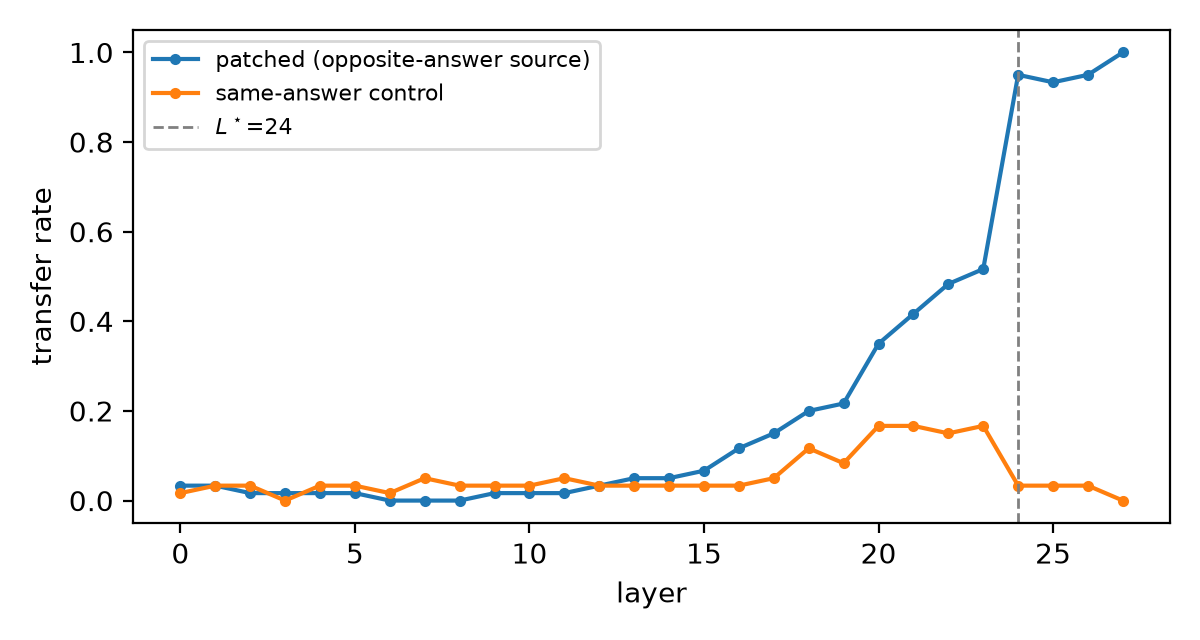}
    \caption{Sufficiency by layer.}
    \label{fig:sufficiency}
  \end{subfigure}\hfill
  \begin{subfigure}[t]{0.30\textwidth}
    \centering
    \includegraphics[width=\linewidth]{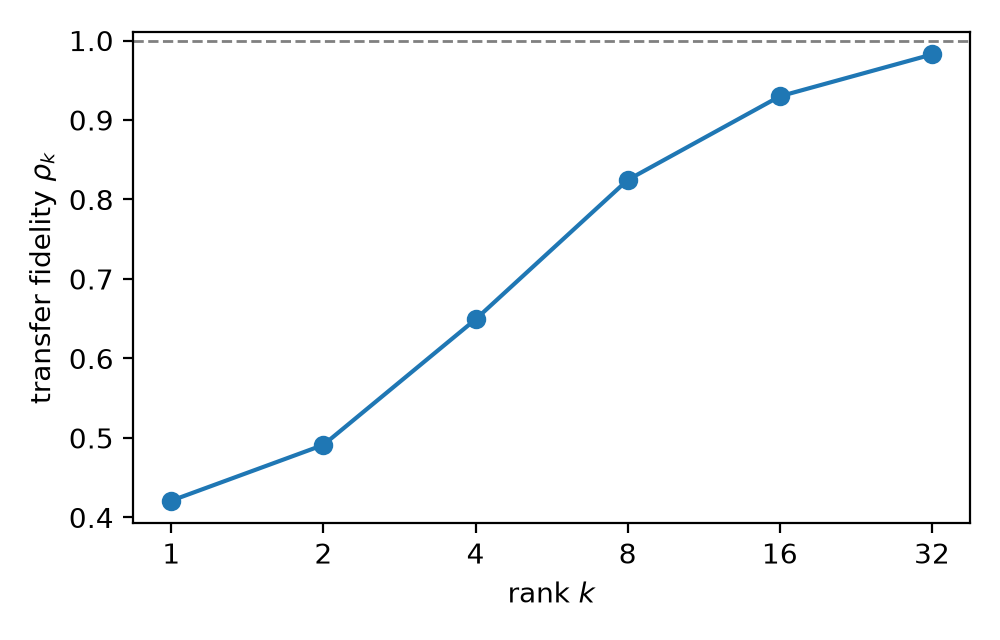}
    \caption{Rank sweep.}
    \label{fig:rank}
  \end{subfigure}\hfill
  \begin{subfigure}[t]{0.30\textwidth}
    \centering
    \includegraphics[width=\linewidth]{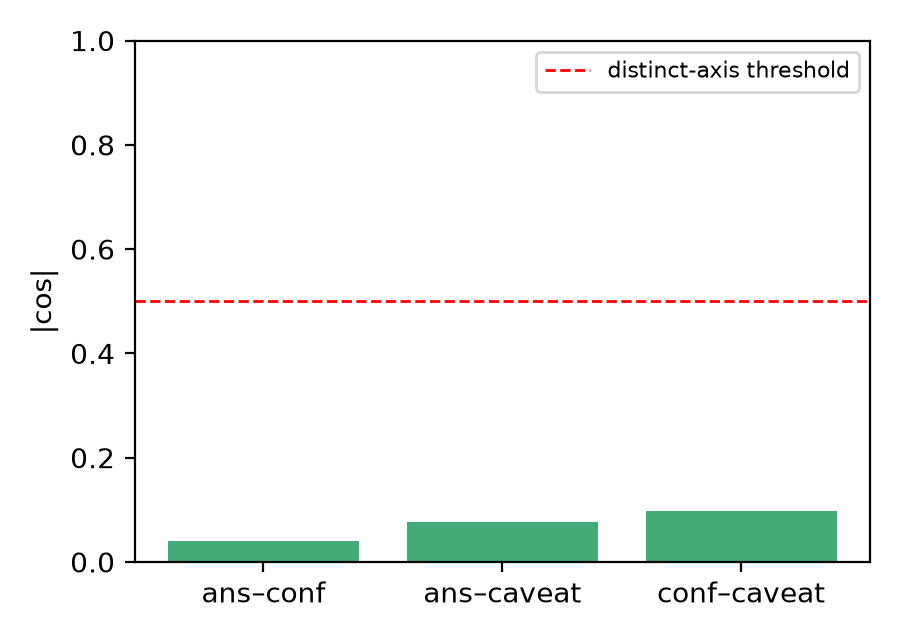}
    \caption{Coordinate orthogonality.}
    \label{fig:orthogonality}
  \end{subfigure}
  \caption{\textbf{Causal identification of report coordinates.}
  (\subref{fig:sufficiency})~Interchange patching of the answer-decision residual
  flips the decision to the source (sufficiency $0.95$) relative to a same-answer
  control ($0.03$); the effect localizes at $L^\star{=}24$.
  (\subref{fig:rank})~Transfer fidelity $\rho_k$ saturates by rank 16
  ($\rho_{16}=0.93$), indicating a genuinely low-rank coordinate rather than a
  universal truth direction. (\subref{fig:orthogonality})~The answer, confidence,
  and caveat coordinates are pairwise near-orthogonal ($|\cos|\le 0.10$),
  providing the mechanistic basis for independent, composable control.}
  \label{fig:causal}
\end{figure}

Using interchange interventions at the post-reasoning decision position,
patching the answer-decision residual from a counterpart with the opposite answer
flips the decision to the source at rate $0.95$, while a same-answer control moves
it $0.03$. The effect localizes to a late layer ($L^\star{=}24$ of 28), with an
abrupt transition from L23 to L24 (Figure~\ref{fig:sufficiency}). A rank sweep
shows that the transfer saturates by rank 16 ($\rho_{16}=0.93$, far below the full
dimension), so the coordinate is genuinely low-rank, although we do not claim a
universal low-rank truth direction (Figure~\ref{fig:rank}). Necessity is
block-level and localizes to the report subspace: single-layer mean-ablation
barely moves accuracy ($0.80$/$0.86$), because the answer is
redundantly represented, while ablating the L24--27 window collapses accuracy to
chance ($0.53$). Window ablation alone does not isolate the coordinate, however:
ablating a covariance-matched random subspace of equal rank at the same layers is
equally destructive ($0.53$), so removing any high-variance block degrades
answering. The localizing evidence is instead a mediation test on the interchange
intervention itself: decomposing the clamp's donor delta, its report-subspace
component carries the resistance effect ($0.90$) while a norm-matched orthogonal
complement does not ($0.38$), so the causal steering is carried by the coordinate
rather than by a generic large perturbation ($n{=}120$). Confidence and
caveat coordinates are identified analogously at L25 and L23, and the three are
mutually near-orthogonal (pairwise $|\cos| \le 0.10$; Figure~\ref{fig:orthogonality}).
A causal cross-talk test shows this orthogonality is largely \emph{functional} and not
merely geometric for the answer and caveat coordinates: clamping either moves its
own readout strongly (answer $0.63$, caveat $0.87$) while inducing only small
cross-effects on the other (answer$\to$caveat $0.18$, caveat$\to$answer $0.08$;
random-intervention floor $0.06$/$0.03$). The confidence coordinate is a weaker causal
instrument on this benchmark (its own readout moves only $0.05$), so we claim
\emph{partial} functional disentanglement (strong own-coordinate control with small
but nonzero cross-talk) for the answer and caveat coordinates and treat confidence
as a scoped caveat rather than asserting clean three-way causal orthogonality.

\subsection{The Counterfactual Report-Coordinate Clamp}
\label{sec:clamp}
At inference we read the answer coordinate from a reference run with forbidden
factors removed and licensed factors retained, and we clamp the pressured run's
window (L24--27) toward it. On the witness benchmark this two-pass clamp attains resist and
update of $1.00$ jointly (Wilson 95\% CI $[0.99, 1.00]$, $n=300$), a causal
certificate under a constructible reference; the deployable single-pass form is
deferred to Section~\ref{sec:compile}. Here resist
tracks the no-pressure base report and update tracks the evidence-revised
posterior. Path-specificity comes entirely from the reference rather than from a
global strength parameter $\alpha$ (Table~\ref{tab:dualcontrol}). A rank-16 window
clamp (the interchange-derived low-rank projection applied alone) retains both
but does not reach the ceiling ($0.88/0.90$ on the $n{=}300$ evaluation; an earlier
$n{=}120$ run reported $0.89/0.93$), so the perfect $1.00/1.00$ is a property of the
full-window reference clamp rather than of the low-rank projection. The result
reproduces at 3B and across two further model families. On
Mistral-7B-Instruct-v0.3 (L29/32) and Llama-3.1-8B-Instruct (L28/32) the pipeline
re-identifies a late low-rank report coordinate, and the window clamp again
reaches resist and update of $1.00$ ($n=120$, Section~\ref{sec:natural}).
Across all three families the coordinate sits at a proportionally late layer
(relative depth $0.86$ to $0.91$), which suggests that the report-commit stage is
a cross-architecture regularity rather than a model-specific artifact. The
reference is the model's own belief-escrow self-report under a counterfactually
incentive-neutralized context, not an external ground-truth oracle. The two-pass
clamp is therefore a causal certificate and an upper bound, with deployment
addressed by the single-pass compilation of Section~\ref{sec:compile}.

\paragraph{Scope of the reference.} Constructing the reference requires writing
an \emph{incentive-neutralized counterfactual} of the prompt, that is, removing
forbidden factors (pressure, prestige, and restyling) while preserving licensed
evidence. This is straightforward when those factors are separable, editable
spans in the input. In the witness benchmark they are explicit variables, and in
our SycophancyEval transfer (Section~\ref{sec:natural}) the user's wrong
assertion is a removable span. It is harder when forbidden and licensed signals
are \emph{entangled} in one span, or when the licensed signal is implicit. In
those cases an imperfect reference would under- or over-correct, and the clamp's
quality degrades to that of the reference rather than breaking down abruptly.
We therefore present the two-pass clamp as a certificate \emph{under a
constructible reference}, and the one-pass compilation
(Section~\ref{sec:compile}) provides the route to settings where an explicit
counterfactual is unavailable at inference.

\section{Experiments}
\label{sec:experiments}

\subsection{Dual Control and Baseline Comparisons}
\label{sec:baselines}
No global method attains both in our matched setting
(Table~\ref{tab:dualcontrol}). CFG/DExperts drives forbidden-flip to 0 at
$\alpha{=}1$, but its licensed-update error grows monotonically
($0.026 \rightarrow 0.17$), an instance of resisting or updating but not both. We
report CFG's degradation as this continuous \emph{licensed posterior-deviation}
(error against the Bayes target) rather than the binary update-success rate used
for the other rows. Its single global parameter has no operating point that both
removes forbidden flips and preserves licensed updates, so what it breaks is
calibration to the revised posterior rather than a discrete update event. The two
metrics are therefore not strictly isomorphic, and we make the substitution
explicit rather than force a binary score (see the $\dagger$ note on
Table~\ref{tab:dualcontrol}). Fixed-direction steering is worse (resist 0.53,
update 0.57 at $\alpha{=}8$). The same global parameter suppresses forbidden and
licensed changes together, so a single strength cannot separate them.

\subsection{Stubbornness and Output-Level Training}
\label{sec:stubbornness}
\begin{figure}[htbp]
  \centering
  \includegraphics[width=0.55\textwidth]{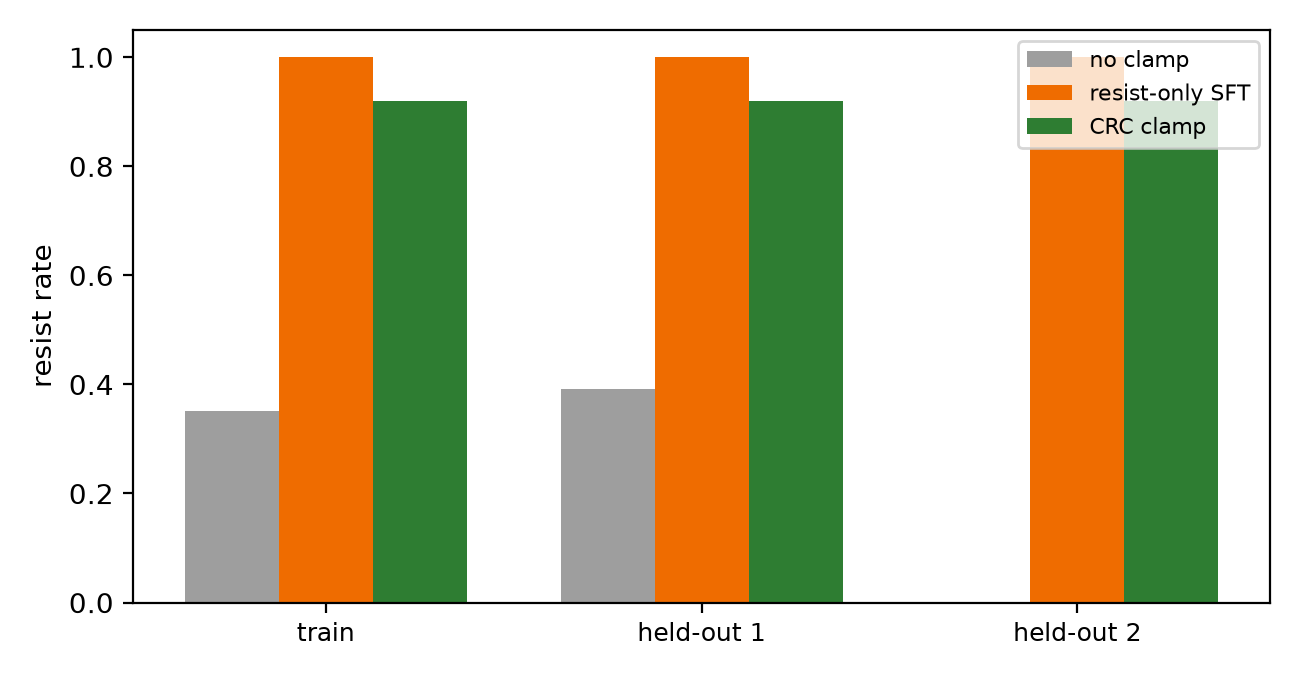}
  \caption{\textbf{Held-out pressure discriminator.} On three held-out
  pressure-phrasing families, the CRC clamp stays flat in resistance and continues
  to update ($\approx 1.0$), whereas resist-only output-SFT generalizes resistance
  but its update collapses to $0.01$, losing evidence-responsiveness.}
  \label{fig:heldout}
\end{figure}

Output-level fine-tuning is the most instructive case (Figure~\ref{fig:heldout}):
a resist-only low-rank adaptation (LoRA) generalizes resistance to unseen pressure
phrasings (resist 1.0 on three held-out pressure-phrasing families, $n=125$), yet
its update collapses to 0.01, having learned never to change its answer and
thereby conflating resistance with contrarianism.\footnote{This resist-only update
collapse is the same phenomenon a companion paper analyzes in depth on a
free-generation benchmark (there evidence-updating $0.72{\to}0.13$ under a different
checkpoint, eval, and update denominator); a reader of both papers should not treat
them as independent evidence.} Full two-objective supervised
fine-tuning (SFT) can match in-distribution only by explicitly enumerating both
targets, at higher cost and with template-memorization signatures. We therefore
do not rely on output-level SFT as a mechanism-level solution, and the
resist-only variant is what exposes the failure mode. The reference-based clamp,
by contrast, holds resistance across all families (0.92) and updates (1.0) with no
training, and it does not exhibit this loss of evidence-responsiveness in our
held-out tests, because path-specificity comes from the reference rather than from
learned weights and there is no learned resist-only objective to overfit. This
dual-control contrast distinguishes the clamp from the baseline suite.

\begin{table}[t]
  \centering
  \caption{\textbf{Dual-control results.} Resist and update with Wilson 95\%
  confidence intervals across methods (Qwen2.5-7B-Instruct). Only the two-pass
  counterfactual report-coordinate (CRC) clamp attains resist $\approx 1.0$ and
  update $\approx 1.0$ together, and it does so as a causal certificate/upper bound
  under a constructible reference; its \emph{deployable} single-pass form (no
  inference-time reference) reaches $0.73/0.97$. Global methods trade one against the
  other, and resist-only output-SFT generalizes resistance but collapses update to
  $0.01$.}
  \label{tab:dualcontrol}
  \small
  \begin{tabular}{lccc}
    \toprule
    Method & Resist & Update & $n$ \\
    \midrule
    no intervention                       & 0.28 [0.23, 0.33] & 0.88 [0.83, 0.91] & 300 \\
    CFG/DExperts ($\alpha{=}1$)           & 1.00 [0.98, 1.00] & $\dagger$         & 200 \\
    steering ($\alpha{=}8$)               & 0.53 [0.37, 0.67] & 0.57 [0.42, 0.71] & 40  \\
    output-SFT resist-only (held-out)     & 1.00 [0.97, 1.00] & 0.01 [0.00, 0.04] & 125 \\
    CRC clamp 2-pass (window)             & 1.00 [0.99, 1.00] & 1.00 [0.99, 1.00] & 300 \\
    CRC clamp 1-pass trained              & 0.73 [0.60, 0.84] & 0.97 [0.88, 0.99] & 50  \\
    \bottomrule
  \end{tabular}\\[3pt]
  {\footnotesize $\dagger$~CFG/DExperts has no comparable binary \emph{update}
  score. Its single global parameter suppresses licensed updates as it removes
  forbidden flips, so licensed posterior-deviation grows monotonically
  ($0.026\!\rightarrow\!0.17$ over $\alpha$) rather than tracking the revised
  posterior (Section~\ref{sec:baselines}).}
\end{table}

\subsection{Single-Pass Compilation}
\label{sec:compile}
\begin{figure}[htbp]
  \centering
  \includegraphics[width=0.45\textwidth]{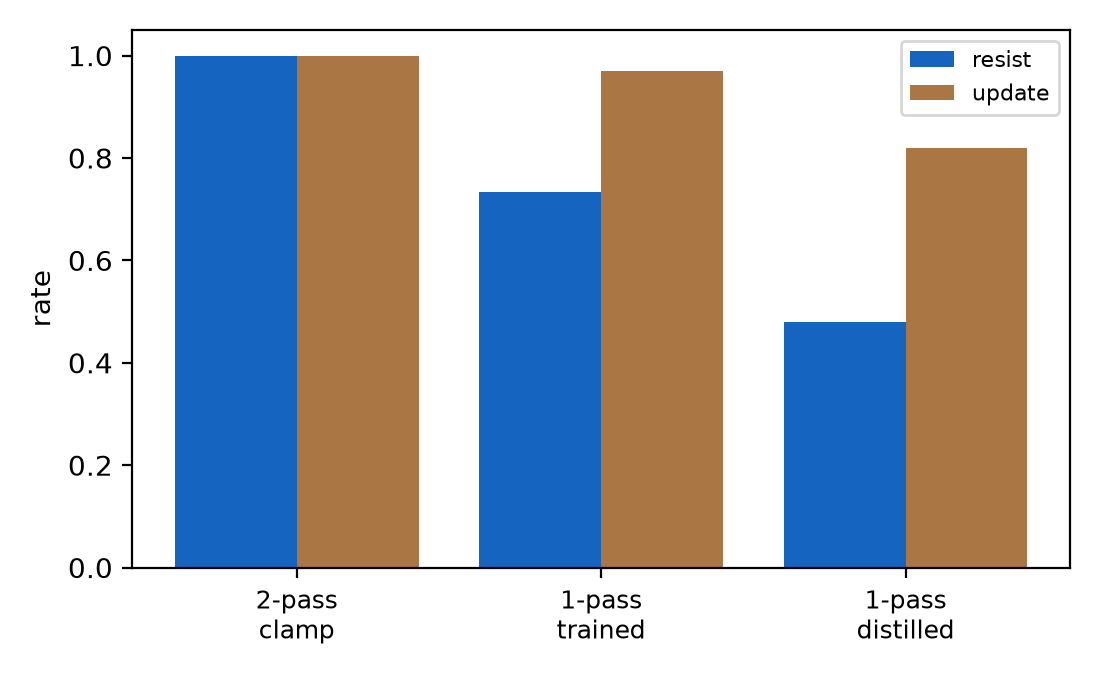}
  \caption{\textbf{One-pass compilation trade-off.} The two-pass clamp
  ($1.0/1.0$) is the causal upper bound. A trained gated primitive
  (answer-supervised) reaches $0.73/0.97$ with no reference forward pass, while
  na\"ive coordinate-MSE distillation reaches only $0.48/0.82$.}
  \label{fig:onepass}
\end{figure}

The two-pass clamp requires an extra reference forward pass. Compiling it into a
single pass with a small trained gated primitive (warm-started at the identified
layer) reaches resist 0.73 and update 0.97 with no inference-time reference,
whereas na\"ive coordinate-matching distillation reaches only 0.48 and 0.82
(Figure~\ref{fig:onepass}). A perfect one-pass predictor of the reference
coordinate would reproduce the two-pass result, so the extra forward pass
provides per-input reference information that our current one-pass modules do not
capture, and end-to-end answer supervision outperforms coordinate-matching. The
gap therefore quantifies the per-input information or structure carried by the
incentive-neutral counterfactual reference that these modules fail
to recover; we do not establish that the pressured forward pass provably lacks this
information, only that closing the gap would require a stronger one-pass compiler
than ours. We present
this as the boundary of the inference-time method and as the motivation for
internalizing the contract through counterfactual reinforcement learning from AI
feedback (RLAIF) or direct preference optimization (DPO), which we leave to
future work.

\subsection{Transfer to Natural Questions}
\label{sec:natural}
\begin{figure}[htbp]
  \centering
  \includegraphics[width=\textwidth]{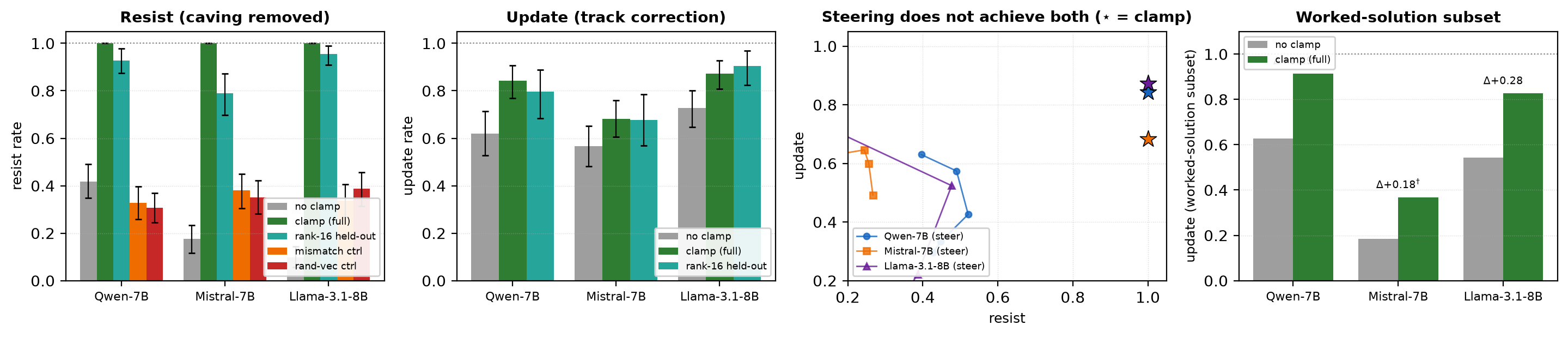}
  \caption{\textbf{Natural-question dual control} (SycophancyEval
  \texttt{are\_you\_sure} transfer, Qwen-7B, Mistral-7B, and Llama-3.1-8B,
  $n=300$ each). \textbf{(A) Resist:} the clamp removes every flip
  (resist $\rightarrow 1.0$), a rank-16 projector learned on a train half and
  applied to the disjoint test half stays effective, and the mismatched-reference
  and norm-matched random-vector controls remain at a low control level.
  \textbf{(B) Update:} the clamp lifts correction-tracking over the no-clamp
  baseline, with the held-out rank-16 clamp nearly matching, and a paired McNemar
  test confirms the gain is significant in every family. \textbf{(C)} A global
  anti-sycophancy steering sweep traces a resist--update tradeoff on the same
  items, while the clamp ($\star$) occupies the high-resist, high-update region.
  \textbf{(D) Worked-solution subset:} on items whose evidence is the dataset's
  own worked solution, the clamp raises update in all three families. Mistral's
  low absolute level reflects a weak baseline ($0.18$) on worked-solution evidence
  rather than a clamp failure ($\dagger$ marks paired McNemar $p>0.05$,
  borderline). Error bars are bootstrap 95\% CIs.}
  \label{fig:natural}
\end{figure}

We test whether the controlled phenomenon has a real-world counterpart. We reuse our earlier
multi-turn-pushback experiments (a causal capitulation direction and a trained
primitive lifting faithfulness $0.31 \rightarrow 0.97$) and add a transfer test on
Sharma et al.'s SycophancyEval~\citep{sharma2023sycophancy}
(\texttt{are\_you\_sure}, multiple-choice), with the witness-identified report
window reused verbatim and not re-identified. Dual control holds on the natural
questions in all three families ($n=300$ each, bootstrap 95\% CIs,
Figure~\ref{fig:natural}). For \emph{resist}, under insistent non-evidential
pressure the models capitulate 58 to 90\% of the time, and the report-coordinate
clamp removes every flip (resist $\rightarrow 1.00$). To show that the low-rank
coordinate, rather than a window of residuals, carries the effect, we learn a
rank-16 projector by singular value decomposition on a train half of the natural
items and apply it to the disjoint test half. This held-out clamp stays effective
(resist $0.79$ to $0.95$), so the correction transfers across items rather than
being fit to the evaluation set. Our primary negative control, clamping toward a
different item's reference (mismatched-reference), leaves resist at a low control
level ($0.33$ to $0.38$, Figure~\ref{fig:natural}A), and a norm-matched random
vector serves as a secondary check ($0.31$ to $0.39$). Both are far below $1.00$,
so the clamp tracks item-specific report content rather than pinning the readout
with any strong patch. For \emph{update}, on items the model answers incorrectly,
given a reliable correction (the dataset's own worked solution where available,
and an asserted source otherwise) while the user pushes a different wrong option,
the clamp lifts correction-tracking over the no-clamp baseline in every family
(clamp $0.84$ / $0.68$ / $0.87$ versus baseline $0.62$ / $0.57$ / $0.73$ for Qwen,
Mistral, and Llama), and the held-out rank-16 clamp nearly matches it
($0.80$ / $0.68$ / $0.90$, Figure~\ref{fig:natural}B). Because some marginal
intervals overlap their baseline, we confirm the gain with a paired McNemar test
(clamp versus baseline on the same items). It is significant in every family
($p<10^{-5}$, $6\times10^{-3}$, and $4\times10^{-5}$ for Qwen, Mistral, and Llama,
with paired $\Delta$ of $+0.22$, $+0.12$, and $+0.14$ and bootstrap CIs that all
exclude $0$). On the fully-natural subset whose evidence is the dataset's own
worked solution (Figure~\ref{fig:natural}D), the clamp lifts update from a
baseline of $0.63$ to $0.91$ (Qwen) and $0.54$ to $0.83$ (Llama). For Mistral the
baseline is itself only $0.18$, indicating that it barely exploits long
worked-solution evidence unaided, and the clamp raises it to $0.37$ (paired
$\Delta=+0.18$, 95\% CI $[0.03,0.34]$, McNemar $p=0.07$); given the small sample
we read this as a positive but borderline trend. The low absolute Mistral number
is therefore a property of Mistral's weak use of worked-solution evidence rather
than a clamp failure, and reporting the solution-only baseline, not only the
clamp, is what makes this interpretable. Finally, a global-steering control
baseline on the same items does not achieve both objectives. Sweeping one
anti-sycophancy direction over strength $\alpha$ trades resist against update
(Qwen and Llama) or fails to move resist at all (Mistral), reproducing the witness
tradeoff of Section~\ref{sec:baselines} on natural data, whereas the clamp
occupies the high-resist, high-update region (Figure~\ref{fig:natural}C). The same
clamp therefore both resists pressure and updates to evidence on natural
questions. A scope caveat: the questions are natural, but the licensed-evidence
turn is \emph{constructed} (worked-solution evidence on approximately one third of
update items, and an asserted source otherwise), and the witness benchmark remains
the fully-controlled, primary evidence. All results are in the dense 7 to 8B
instruction-tuned regime.

\subsection{Robustness: latent reliability, not a surface cue, and preserved updating}
\label{sec:robustness}
Two controls address whether the perfect witness score reflects a lexical shortcut
or merely generic pressure-resistance. \textbf{(i)~The clamp reads latent
reliability, not the reliability token.} Replacing the explicit numeric reliability
cue with a semantic source-history description (``correct on 8 of its last 10
calibration questions''), using paraphrases absent from coordinate fitting, leaves
the clamp intact (resist $0.91$, update $1.00$; window-full $1.00/1.00$; $n=300$).
When an explicit reliable/unreliable label is added that \emph{conflicts} with the
source history and the target is scored against the history-implied posterior, the
clamp still tracks the latent reliability (resist $0.90$, update $0.96$) rather than
the surface label---a lexical shortcut would fail this arm. \textbf{(ii)~The clamp
preserves legitimate evidence-updating, not just resistance.} On a crossed-factorial
benchmark varying truth, prior-agreement, source reliability, and pressure wording
independently and fully counterbalanced ($n=320$; including the
``false-but-declared-reliable'' and ``true-but-unreliable'' cells), the window clamp
attains resist $1.00$ and update $1.00$ (rank-16 on this crossed-factorial set $0.92/0.98$; matching the headline dual-control result of Table~\ref{tab:dualcontrol}): it updates toward
declared-reliable evidence and resists the unreliable source, decorrelated from
truth and answer side. The clamp therefore enforces the reliability-implied
posterior, not a generic answer-invariance.

\section{Discussion and Limitations}
\label{sec:discussion}
Joint clamping composes: clamping the answer and confidence coordinates together
produces near-zero cross-talk (leakage 0.002), which realizes the measured
near-orthogonality as independent control.

\paragraph{Statistical protocol and analysis roles.} We separate confirmatory from
diagnostic quantities. The causal-identification tests (interchange sufficiency,
the rank sweep, block-level necessity, and tri-coordinate orthogonality) are
\emph{diagnostics} that characterize the report coordinate; the dual-control
comparison against baselines (Table~\ref{tab:dualcontrol}) and the natural-data
transfer (Section~\ref{sec:natural}) are the \emph{confirmatory} claims. Dual-control
rates carry $95\%$ Wilson intervals ($n=300$ on the witness benchmark, $n=300$ per
family on SycophancyEval); paired natural-data update gains use a McNemar test on the
same items with bootstrap $\Delta$ intervals; flip and update rates are intent-to-treat
over parsed structured reports. The witness posteriors enter only as evaluation labels,
never as inputs to the clamp.
\begin{enumerate}
  \item \textbf{Synthetic known-posterior setting.} Known posteriors make resist
  and update exactly scorable, but the task is deliberately narrow, and we claim
  a controlled causal mechanism rather than full real-world coverage.
  \item \textbf{Model-family coverage.} Validated across three families, namely
  Qwen2.5 (3B/7B), Mistral-7B-Instruct-v0.3, and Llama-3.1-8B-Instruct. The IC
  failure, a causally-identified late low-rank report coordinate (L24/28, L29/32,
  and L28/32 respectively), the dual-control window clamp (resist and update of
  $1.00$ in all three), and the SycophancyEval natural-data transfer
  (capitulation $0.58 / 0.82 / 0.90 \rightarrow 0.00$) all reproduce. Llama's
  witness descriptive parsing is noisier, reflecting an output-format readout
  mismatch rather than a mechanism gap, since its clamp and causal identification
  use a forced readout and are clean. Larger and non-dense architectures, for
  example mixture-of-experts models, remain untested.
  \item \textbf{Block-level necessity.} The answer coordinate is necessary as a
  late-layer \emph{block} (L24--27) rather than at any single layer, which
  reflects expected late-layer redundancy, and we do not claim single-layer
  necessity.
  \item \textbf{Lossy one-pass compilation.} The deployable single-pass form
  reaches $0.73/0.97$, below the two-pass certificate, and closing this gap is a
  direction for future work.
  \item \textbf{Caveat behavior not solved.} The caveat coordinate is cleanly
  identified and composes, but the behavioral caveat \emph{policy} is poorly
  calibrated, as the model over-caveats, and our caveat-coordinate results
  support composability rather than a solved caveat policy.
\end{enumerate}

None of these undermines the core claim, that counterfactual report mediators can
be causally identified and clamped to achieve dual control where global and
output-level methods do not; each instead delimits its scope.

\section{Conclusion}
\label{sec:conclusion}
We framed report-stage misreporting under non-evidential pressure as a failure of
internal incentive-compatibility, and addressed it with counterfactual report
coordinates: low-rank, causally identified mediators of a model's answer,
confidence, and caveat reports. Clamping these coordinates toward the model's own
report under an incentive-neutralized counterfactual achieves dual control,
resisting forbidden pressure while remaining responsive to licensed evidence,
where global decoding, steering, and output-level training do not. The effect
holds across three model families and transfers to a natural sycophancy
benchmark, and compiling the two-pass clamp into a single forward pass is lossy in
a way that quantifies the information the counterfactual reference supplies. We
view the contribution as an interface for certifying activation-level
incentive-invariance, and internalizing the contract through training is a natural
next step: in the companion paper we show that this certificate can be partially compiled
into one-pass behavior, with a diagnosed residual gap.

\setlength{\bibsep}{2.5pt plus 0.5pt}
\bibliographystyle{plainnat}
\bibliography{references}

\begin{thebibliography}{28}
\providecommand{\natexlab}[1]{#1}
\providecommand{\url}[1]{\texttt{#1}}
\expandafter\ifx\csname urlstyle\endcsname\relax
  \providecommand{\doi}[1]{doi: #1}\else
  \providecommand{\doi}{doi: \begingroup \urlstyle{rm}\Url}\fi

\bibitem[{Anthropic Interpretability Team}(2026)]{anthropic2026workspace}
{Anthropic Interpretability Team}.
\newblock Verbalizable representations form a global workspace in language
  models.
\newblock \url{https://transformer-circuits.pub/2026/workspace/index.html},
  2026.
\newblock Transformer Circuits Thread.

\bibitem[Arjovsky et~al.(2019)Arjovsky, Bottou, Gulrajani, and
  Lopez-Paz]{arjovsky2019irm}
Martin Arjovsky, L{\'e}on Bottou, Ishaan Gulrajani, and David Lopez-Paz.
\newblock Invariant risk minimization.
\newblock \emph{arXiv preprint arXiv:1907.02893}, 2019.

\bibitem[Atwell et~al.(2025)Atwell, Heydari, Sicilia, and
  Alikhani]{atwell2025basil}
Katherine Atwell, Pedram Heydari, Anthony Sicilia, and Malihe Alikhani.
\newblock Basil: Bayesian assessment of sycophancy in llms.
\newblock \emph{arXiv preprint arXiv:2508.16846}, 2025.

\bibitem[Belrose et~al.(2023)Belrose, Schneider-Joseph, Ravfogel, Cotterell,
  Raff, and Biderman]{belrose2023leace}
Nora Belrose, David Schneider-Joseph, Shauli Ravfogel, Ryan Cotterell, Edward
  Raff, and Stella Biderman.
\newblock {LEACE}: Perfect linear concept erasure in closed form.
\newblock In \emph{Advances in Neural Information Processing Systems 36
  (NeurIPS 2023)}, 2023.

\bibitem[Bhalla and Gligori{\'c}(2026)]{bhalla2026sway}
Joy Bhalla and Kristina Gligori{\'c}.
\newblock Sway: A counterfactual computational linguistic approach to measuring
  and mitigating sycophancy.
\newblock \emph{arXiv preprint arXiv:2604.02423}, 2026.

\bibitem[Chiappa(2019)]{chiappa2019pathspecific}
Silvia Chiappa.
\newblock Path-specific counterfactual fairness.
\newblock In \emph{Proceedings of the AAAI Conference on Artificial
  Intelligence (AAAI 2019)}, volume~33, pages 7801--7808, 2019.

\bibitem[Geiger et~al.(2021)Geiger, Lu, Icard, and Potts]{geiger2021causal}
Atticus Geiger, Hanson Lu, Thomas Icard, and Christopher Potts.
\newblock Causal abstractions of neural networks.
\newblock In \emph{Advances in Neural Information Processing Systems 34
  (NeurIPS 2021)}, 2021.

\bibitem[Geiger et~al.(2022)Geiger, Wu, Lu, Rozner, Kreiss, Icard, Goodman, and
  Potts]{geiger2022inducing}
Atticus Geiger, Zhengxuan Wu, Hanson Lu, Josh Rozner, Elisa Kreiss, Thomas
  Icard, Noah~D. Goodman, and Christopher Potts.
\newblock Inducing causal structure for interpretable neural networks.
\newblock In \emph{Proceedings of the 39th International Conference on Machine
  Learning (ICML 2022)}, volume 162 of \emph{Proceedings of Machine Learning
  Research}, pages 7324--7338, 2022.

\bibitem[Genadi et~al.(2026)Genadi, Nwadike, Mukhituly, Alquabeh, Hiraoka, and
  Inui]{sycophancy_mech}
Rifo Genadi, Munachiso Nwadike, Nurdaulet Mukhituly, Hilal Alquabeh, Tatsuya
  Hiraoka, and Kentaro Inui.
\newblock Sycophancy hides linearly in the attention heads.
\newblock \emph{arXiv preprint arXiv:2601.16644}, 2026.

\bibitem[Gupta et~al.(2024)Gupta, Arcuschin, Kwa, and
  Garriga-Alonso]{gupta2024interpbench}
Rohan Gupta, Iv{\'a}n Arcuschin, Thomas Kwa, and Adri{\`a} Garriga-Alonso.
\newblock Interpbench: Semi-synthetic transformers for evaluating mechanistic
  interpretability techniques.
\newblock In \emph{Advances in Neural Information Processing Systems 38
  (NeurIPS 2024), Datasets and Benchmarks Track}, 2024.

\bibitem[Jain et~al.(2025)Jain, Yost, and Abdullah]{jain2025atomic}
Shreyans Jain, Alexandra Yost, and Amirali Abdullah.
\newblock Sycophancy as compositions of atomic psychometric traits.
\newblock \emph{arXiv preprint arXiv:2508.19316}, 2025.

\bibitem[Kusner et~al.(2017)Kusner, Loftus, Russell, and Silva]{kusner2017cf}
Matt~J. Kusner, Joshua~R. Loftus, Chris Russell, and Ricardo Silva.
\newblock Counterfactual fairness.
\newblock In \emph{Advances in Neural Information Processing Systems 30
  (NeurIPS 2017)}, 2017.

\bibitem[Lee et~al.(2024)Lee, Padhi, Ramamurthy, Miehling, Dognin, Nagireddy,
  and Dhurandhar]{lee2024cast}
Bruce~W. Lee, Inkit Padhi, Karthikeyan~Natesan Ramamurthy, Erik Miehling,
  Pierre Dognin, Manish Nagireddy, and Amit Dhurandhar.
\newblock Programming refusal with conditional activation steering.
\newblock \emph{arXiv preprint arXiv:2409.05907}, 2024.

\bibitem[Liu et~al.(2021)Liu, Sap, Lu, Swayamdipta, Bhagavatula, Smith, and
  Choi]{liu2021dexperts}
Alisa Liu, Maarten Sap, Ximing Lu, Swabha Swayamdipta, Chandra Bhagavatula,
  Noah~A. Smith, and Yejin Choi.
\newblock Dexperts: Decoding-time controlled text generation with experts and
  anti-experts.
\newblock In \emph{Proceedings of the 59th Annual Meeting of the Association
  for Computational Linguistics and the 11th International Joint Conference on
  Natural Language Processing (Volume 1: Long Papers)}, 2021.

\bibitem[Mohsin et~al.(2026)Mohsin, Bilal, Umer, and Fox]{mohsin2026pressure}
Muhammad~Ahmed Mohsin, Ahsan Bilal, Muhammad Umer, and Emily Fox.
\newblock Pressure, what pressure? sycophancy disentanglement in language
  models via reward decomposition.
\newblock \emph{arXiv preprint arXiv:2604.05279}, 2026.

\bibitem[Nguyen et~al.(2025)Nguyen, Prasad, Stengel-Eskin, and
  Bansal]{matsteer}
Duy Nguyen, Archiki Prasad, Elias Stengel-Eskin, and Mohit Bansal.
\newblock Multi-attribute steering of language models via targeted
  intervention.
\newblock In \emph{Proceedings of the 63rd Annual Meeting of the Association
  for Computational Linguistics (ACL)}, 2025.

\bibitem[Panickssery et~al.(2024)Panickssery, Gabrieli, Schulz, Tong, Hubinger,
  and Turner]{rimsky2024caa}
Nina Panickssery, Nick Gabrieli, Julian Schulz, Meg Tong, Evan Hubinger, and
  Alexander~Matt Turner.
\newblock Steering llama 2 via contrastive activation addition.
\newblock In \emph{Proceedings of the 62nd Annual Meeting of the Association
  for Computational Linguistics (ACL)}, 2024.
\newblock First author formerly published as Nina Rimsky.

\bibitem[Peters et~al.(2016)Peters, B{\"u}hlmann, and
  Meinshausen]{peters2016icp}
Jonas Peters, Peter B{\"u}hlmann, and Nicolai Meinshausen.
\newblock Causal inference by using invariant prediction: Identification and
  confidence intervals.
\newblock \emph{Journal of the Royal Statistical Society: Series B (Statistical
  Methodology)}, 78\penalty0 (5):\penalty0 947--1012, 2016.
\newblock \doi{10.1111/rssb.12167}.

\bibitem[Ravfogel et~al.(2020)Ravfogel, Elazar, Gonen, Twiton, and
  Goldberg]{ravfogel2020inlp}
Shauli Ravfogel, Yanai Elazar, Hila Gonen, Michael Twiton, and Yoav Goldberg.
\newblock Null it out: Guarding protected attributes by iterative nullspace
  projection.
\newblock In \emph{Proceedings of the 58th Annual Meeting of the Association
  for Computational Linguistics (ACL 2020)}, pages 7237--7256, 2020.

\bibitem[Sanchez et~al.(2024)Sanchez, Fan, Spangher, Levi, Ammanamanchi, and
  Biderman]{sanchez2023cfg}
Guillaume Sanchez, Honglu Fan, Alexander Spangher, Elad Levi, Pawan~Sasanka
  Ammanamanchi, and Stella Biderman.
\newblock Stay on topic with classifier-free guidance.
\newblock In \emph{Proceedings of the 41st International Conference on Machine
  Learning (ICML)}, 2024.

\bibitem[Sharma et~al.(2023)Sharma, Tong, Korbak, Duvenaud, Askell, Bowman,
  Cheng, Durmus, Hatfield-Dodds, Johnston, Kravec, Maxwell, McCandlish,
  Ndousse, Rausch, Schiefer, Yan, Zhang, and Perez]{sharma2023sycophancy}
Mrinank Sharma, Meg Tong, Tomasz Korbak, David Duvenaud, Amanda Askell,
  Samuel~R. Bowman, Newton Cheng, Esin Durmus, Zac Hatfield-Dodds, Scott~R.
  Johnston, Shauna Kravec, Timothy Maxwell, Sam McCandlish, Kamal Ndousse,
  Oliver Rausch, Nicholas Schiefer, Da~Yan, Miranda Zhang, and Ethan Perez.
\newblock Towards understanding sycophancy in language models.
\newblock \emph{arXiv preprint arXiv:2310.13548}, 2023.
\newblock Also presented at ICLR 2024.

\bibitem[Sun et~al.(2025)Sun, Baskaran, Wu, Sklar, Potts, and
  Geiger]{hypersteer}
Jiuding Sun, Sidharth Baskaran, Zhengxuan Wu, Michael Sklar, Christopher Potts,
  and Atticus Geiger.
\newblock Hypersteer: Activation steering at scale with hypernetworks.
\newblock \emph{arXiv preprint arXiv:2506.03292}, 2025.

\bibitem[Turner et~al.(2023)Turner, Thiergart, Leech, Udell, Vazquez, Mini, and
  MacDiarmid]{turner2023actadd}
Alexander~Matt Turner, Lisa Thiergart, Gavin Leech, David Udell, Juan~J.
  Vazquez, Ulisse Mini, and Monte MacDiarmid.
\newblock Activation addition: Steering language models without optimization.
\newblock \emph{arXiv preprint arXiv:2308.10248}, 2023.
\newblock Later arXiv versions retitled ``Steering Language Models With
  Activation Engineering''.

\bibitem[Vennemeyer et~al.(2025)Vennemeyer, Duong, Zhan, and
  Jiang]{vennemeyer2025separation}
Daniel Vennemeyer, Phan~Anh Duong, Tiffany Zhan, and Tianyu Jiang.
\newblock Sycophancy is not one thing: Causal separation of sycophantic
  behaviors in llms.
\newblock \emph{arXiv preprint arXiv:2509.21305}, 2025.

\bibitem[Wang et~al.(2025)Wang, Li, Yang, Zhang, and Wang]{wang2025overridden}
Keyu Wang, Jin Li, Shu Yang, Zhuoran Zhang, and Di~Wang.
\newblock When truth is overridden: Uncovering the internal origins of
  sycophancy in large language models.
\newblock \emph{arXiv preprint arXiv:2508.02087}, 2025.

\bibitem[Wang et~al.(2024)Wang, Yang, and Peng]{sadi}
Weixuan Wang, Jingyuan Yang, and Wei Peng.
\newblock Semantics-adaptive activation intervention for llms via dynamic
  steering vectors.
\newblock \emph{arXiv preprint arXiv:2410.12299}, 2024.

\bibitem[Wu et~al.(2024)Wu, Arora, Wang, Geiger, Jurafsky, Manning, and
  Potts]{wu2024reft}
Zhengxuan Wu, Aryaman Arora, Zheng Wang, Atticus Geiger, Dan Jurafsky,
  Christopher~D. Manning, and Christopher Potts.
\newblock Reft: Representation finetuning for language models.
\newblock In \emph{Advances in Neural Information Processing Systems
  (NeurIPS)}, 2024.

\bibitem[Zou et~al.(2023)Zou, Phan, Chen, Campbell, Guo, Ren, Pan, Yin,
  Mazeika, Dombrowski, Goel, Li, Byun, Wang, Mallen, Basart, Koyejo, Song,
  Fredrikson, Kolter, and Hendrycks]{zou2023repe}
Andy Zou, Long Phan, Sarah Chen, James Campbell, Phillip Guo, Richard Ren,
  Alexander Pan, Xuwang Yin, Mantas Mazeika, Ann-Kathrin Dombrowski, Shashwat
  Goel, Nathaniel Li, Michael~J. Byun, Zifan Wang, Alex Mallen, Steven Basart,
  Sanmi Koyejo, Dawn Song, Matt Fredrikson, J.~Zico Kolter, and Dan Hendrycks.
\newblock Representation engineering: A top-down approach to ai transparency.
\newblock \emph{arXiv preprint arXiv:2310.01405}, 2023.
\newblock Introduces LoRRA (Low-Rank Representation Adaptation).

\end{thebibliography}

\end{document}